\pdfoutput=1

\documentclass[11pt]{article}

\usepackage[]{EMNLP2022}

\usepackage{times}
\usepackage{latexsym}

\usepackage[T1]{fontenc}

\usepackage[utf8]{inputenc}

\usepackage{microtype}

\usepackage[scaled]{beramono}

\usepackage{xcolor}
\usepackage[T1]{fontenc}
\usepackage{color-edits}
\usepackage{textcomp}
\usepackage{listings}
\usepackage{lipsum}

\addauthor{MM}{purple} 

\definecolor{codegreen}{rgb}{0,0.6,0}
\definecolor{codegray}{rgb}{0.5,0.5,0.5}
\definecolor{codepurple}{rgb}{0.58,0,0.82}
\definecolor{backcolour}{rgb}{0.95,0.95,0.92}

\lstdefinestyle{code_style}{
    backgroundcolor=\color{backcolour},   
    commentstyle=\color{codegreen},
    keywordstyle=\color{magenta},
    numberstyle=\tiny\color{codegray},
    numbers=none,
    upquote=true,
    stringstyle=\color{codepurple},
    basicstyle=\ttfamily~\footnotesize,
    breakatwhitespace=false,         
    breaklines=false,                 
    captionpos=b,                    
    keepspaces=true,                 
    numbers=left,                    
    numbersep=5pt,                  
    showspaces=false,                
    showstringspaces=false,
    showtabs=false,                  
    tabsize=2,
}

\usepackage{graphicx,scalerel,xparse}

\NewDocumentCommand\hfemoji{}{
    \scalerel*{
        \includegraphics{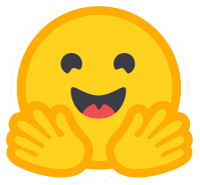}
    }{X}
}
\title{\hfemoji Evaluate \& Evaluation on the Hub:\\Better Best Practices for Data and Model Measurements}

\author{Leandro von Werra\thanks{~~Equal contribution.}, Lewis Tunstall~\footnotemark[1], Abhishek Thakur~\footnotemark[1], Alexandra Sasha Luccioni~\footnotemark[1],\\
{\bf Tristan Thrush, Aleksandra Piktus, Felix Marty, Nazneen Rajani,}\\
{\bf Victor Mustar, Helen Ngo, Omar Sanseviero, Mario Šaško, } \\
{\bf Albert Villanova, Quentin Lhoest, Julien Chaumond, } \\
{\bf Margaret Mitchell, Alexander M. Rush, Thomas Wolf, Douwe Kiela
}\\
  Hugging Face, Inc.\\
  \texttt{ \{leandro,lewis,abhishek,sasha.luccioni,douwe\}@huggingface.co} \\}

\begin{document}
\maketitle
\begin{abstract}
Evaluation is a key part of machine learning (ML), yet there is a lack of support and tooling to enable its informed and systematic practice. 
We introduce~\hfemoji \emph{Evaluate} and \emph{Evaluation on the Hub}---a set of tools to facilitate the evaluation of models and datasets in ML.~\hfemoji  \emph{Evaluate} is a library to support best practices for measurements, metrics, and comparisons of data and models. Its goal is to support reproducibility of evaluation, centralize and document the evaluation process, and broaden evaluation to cover more facets of model performance. It includes over 50 efficient canonical implementations for a variety of domains and scenarios, interactive documentation, and the ability to easily share implementations and outcomes. The library is available at \href{https://github.com/huggingface/evaluate}{https://github.com/huggingface/evaluate}. In addition, we introduce \emph{Evaluation on the Hub}, a platform that enables the large-scale evaluation of over 75,000 models and 11,000 datasets on the Hugging Face Hub, for free, at the click of a button. Evaluation on the Hub is available at \href{https://huggingface.co/autoevaluate}{https://huggingface.co/autoevaluate}.

\textbf{Demo screencast:} \href{https://youtu.be/6rU177zRj8Q}{youtu.be/6rU177zRj8Q}

\end{abstract}


\section{Introduction}
\label{sec:introduction}

Evaluation is a crucial cornerstone of machine learning---not only can it help us gauge whether and how much progress we are making as a field, it can also help determine which model is most suitable for deployment in a given use case. 
However, while the progress made in terms of hardware and algorithms  might look incredible to a ML practitioner from several decades ago, the way we evaluate models has changed very little. 
In fact, there is an emerging consensus that in order to meaningfully track progress in our field, we need to address serious issues in the way in which we evaluate ML systems~\cite{kiela2021dynabench,bowman2021will,raji2021ai,hutchinson2022evaluation}. 

\begin{figure}[t]
    \centering
    \includegraphics[width=0.5\textwidth]{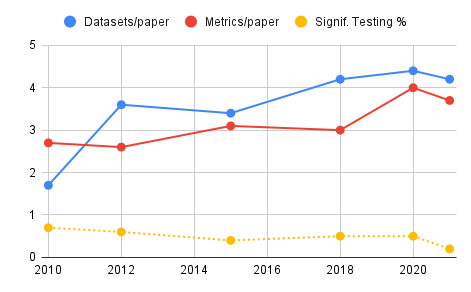}
    \caption{Average number of evaluation datasets and metrics per paper, based on 10 random samples per year from EMNLP proceedings over the past two decades. More recent papers use more datasets and metrics, while fewer of them report statistical significance test results.}
    \label{fig:emnlp}
\end{figure}

In order to have a clearer idea regarding the way model evaluation has evolved in our field, we have carried out our own analysis on a random sample of EMNLP papers from the past two decades, and present our results in Figure~\ref{fig:emnlp}. It can be observed that the number of evaluation datasets and metrics per paper has increased over time, suggesting that model evaluation is becoming increasingly complex and heterogeneous. However, auxiliary techniques such as testing for significance, measuring statistical power, and using appropriate sampling methods have become less common, making results harder to judge when comparing new results to previous work.
We believe that while datasets are now more easily accessible thanks to shared repositories~\cite{lhoest2021datasets}, model evaluation is still unnecessarily cumbersome, with a fragmented ecosystem and a lack of consensus around evaluation approaches and best practices. 

The goal of this work is to address three practical challenges in model evaluation for ML: reproducibility, centralization, and coverage. 

\textit{Reproducibility}: ML systems are extremely sensitive to small (and often undocumented) choices such as random seeds and hyperparameters~\cite{pineau2021improving}. Model performance is often not compared with proper statistical testing that takes this variance into account, making many self-reported comparisons unreliable. Our goal is to standardize this process and thereby improve the reproduction of ML evaluations.

\textit{Centralization}: Historically, ML metrics have been poorly documented, exacerbating an already insufficient community-wide understanding of their usage and shortcomings~\cite{post2018call}.  As metrics and datasets change, the onus is on the community to keep results up-to-date, causing unnecessary replication of work~\cite{ma2021dynaboard} and the proliferation of outdated artifacts~\cite{luccioni2022framework}.
 
\textit{Coverage}: ML as a field still focuses heavily on accuracy-based metrics.
While important, this focus glosses over other critical facets such as  efficiency~\cite{efficientqa2021}, bias and fairness~\cite{qian2022perturbation}, robustness~\cite{goel2021robustness}, and how these factor into choosing a model~\cite{ethayarajh2020utility,ma2021dynaboard}. 
 
 We introduce the open source \emph{Evaluate} library and the \emph{Evaluation on the Hub} platform to address many of these problems. We believe that better evaluation can happen, if we---as a community---establish better best practices and remove hurdles. 

\section{Related work}

\paragraph{Open-Source Tools for Evaluation}  There is a long history of open source projects aiming to capture various measurements, metrics and statistical testing methods for ML. Torchmetrics~\cite{detlefsen22metrics} implements a large number of model evaluation metrics for PyTorch~\cite{paszke2019pytorch}, which is similar to evaluation metrics found in Keras~\cite{chollet2015keras} for TensorFlow. Libraries like Scikit-learn~\cite{pedregosa2011scikit}, SciPy~\cite{virtanen2020scipy}, Statsmodels~\cite{seabold2010statsmodels}, NLTK~\cite{bird2009natural}, TrecTools~\cite{palotti2019}, RL Reliability Metrics~\cite{chan2020rl},  NetworkX~\cite{hagberg2008exploring}, Scikit-image~\cite{van2014scikit}, GEM~\cite{gehrmann-etal-2021-gem}, TorchFidelity~\cite{obukhov2020torchfidelity} also support many evaluation measures across many domains. As integrating metrics into specific frameworks can be difficult, there are also many libraries dedicated to individual evaluations for example rouge\_score,~\footnote{\href{https://github.com/google-research/google-research/tree/master/rouge}{github.com/google-research/google-research/tree/master/rouge}} BARTScore~\cite{yuan2021bartscore}, or SacreBLEU~\cite{post2018call}. The fragmentation of the ecosystem leads to various problems,  such as a wide range of incompatible conventions and APIs, or misreporting due to differing implementations and results. 

In \textit{Evaluate}, we provide one single interface backed by a centralized Hub. Metrics can easily be shared, are version controlled, have a standardized interface, and allow for multimodal inputs.

\paragraph{Evaluation as a Service} The idea of \emph{Evaluation as a Service} \cite{ma2021dynaboard,kiela2021dynabench}, whereby models are submitted for another party to be centrally evaluated, has recently gained traction as a more reproducible way to conduct model evaluation. Central evaluation also facilitates holding challenges and competitions around datasets~\cite{EvalAI,pavao2022codalab,akhbardeh-etal-2021-findings} as opposed to simply evaluating self-reported model results or comparing model scores with benchmark suites \cite{bajaj2016ms,coleman2017dawnbench,wang2018glue,wang2019superglue,kardas-etal-2020-axcell,reddi-2020-mlperf,liu2021explainaboard,goel2021robustness,dror2019deep}. The advantages of conducting evaluation centrally are multiple, including better reproducibility, forward/backward compatibility, and the ability to measure models along multiple axes of evaluation~(e.g. efficiency and fairness, in addition to accuracy), which can help contribute towards a more systematic approach to evaluation.

\paragraph{Issues with Evaluation}
Several studies of ML research and practice have been carried out in recent years on different aspects pertaining to ML evaluation, and together they paint a bleak picture of evaluation in our field. For instance, a 2019 large-scale replication study of 255 ML papers found that only 63\% of the results they reported could be systematically replicated~\cite{raff2019step}. A complementary survey of 3,800 papers from \textit{Papers with Code} has shown that a large majority of metrics used do not adequately reflect models' performance and that they largely do not correlate with human judgement~\cite{Blagec-et-al-2021}. Finally, a recent study of 770 papers in machine translation from the last decade found that while 108 new metrics have been proposed for the task, 99.8\% of papers continue to use BLEU score for reporting results~\cite{marie2021scientific}, despite the fact that the original BLEU score~\cite{papineni2002bleu} has been shown to vary based on user-chosen parameters such as tokenization, which vary across languages~\cite{post2018call,ananthakrishnan2007some}. These issues motivate the development of the tools presented in this work.

\section{Library: \textit{Evaluate}}

The \emph{Evaluate} library provides canonical implementations of a large set of  \textit{evaluation modules}. Modules are available to the community via a single, easy-to-use API. We provide extensive and detailed documentation cards for each, describing their correct usage, range of values and possible pitfalls, in a similar vein to model and dataset cards~\cite{mitchell2019modelcards,gebru2021datasheets}. To facilitate extensibility, each evaluation model lives in a separate Git repository, and new modules can be easily contributed. The core library is released under the Apache 2.0 license and is available on GitHub,~\footnote{\href{https://github.com/huggingface/evaluate}{github.com/huggingface/evaluate}} making it easy to adopt and deploy.

The library is designed to address the main challenges discussed in Section \ref{sec:introduction}. Metrics are versioned and documented to support \textit{reproducibility} within the framework. The core system is \textit{centralized} to facilitate comparisons across models in a consistent manner supporting best practices, and data is stored in Git to allow backups and cloning. Finally, the tool is inherently designed for a multi-model, multi-evaluation paradigm supporting broad evaluation \textit{coverage} by default. 

\subsection{Library Structure}
\emph{Evaluate} aims to support a range of model and dataset comparisons. It offers three distinct types of evaluation modules:

\paragraph{Metrics:} Metrics to provide a score for model performance (e.g. accuracy or BLEU score). They play a central role for decisions around the use and deployment of models, allowing models to be compared and evaluated based on given benchmarks.
\paragraph{Comparisons:} Comparisons are used to compare the predictions of two models (e.g. McNemar's test). When comparing two models, these scores can help determine whether the difference in the models' behavior is statistically significant.
\paragraph{Measurements:} Measurements are used to investigate the characteristics of a dataset (e.g. fraction of duplicates, skew in label distribution). These statistics are a crucial step for gleaning more insights regarding training or evaluation datasets.

\subsection{Library Tour}

We demonstrate how \emph{Evaluate} works with a quick tour of its features. In this section we focus on metrics, but the showcased methods work identically for the other types of evaluation modules.

\paragraph{Core Library} Any metric, measurement, or comparison can be loaded using its name. 
\begin{lstlisting}[language=Python, numbers=none]
 import evaluate
 metric = evaluate.load("accuracy")
\end{lstlisting}
The name can refer to a local file path or the name of a repository on the Hugging Face Hub. 

Users can add predictions and/or references one at a time or pass all of them directly to \texttt{compute()}.

\begin{lstlisting}[language=Python, numbers=none]
 # batches can be added sequentially
 metric.add_batch(predictions = [1, 1],
                  references =  [1, 0])
 metric.compute()

 # or in one compute call
 metric.compute(predictions = [1, 1],
                references  = [1, 0])
\end{lstlisting}

Note that the sequential method is particularly useful in a multi-worker setup, where each worker adds data and the compute operation happens at the end. \emph{Evaluate} uses Apache Arrow as its backend, which means that adding data to the metric does not use any additional memory. The full set of data is only loaded when the metric is computed.

Several metrics can be bundled together and follow the same API as a single metric, returning all results at once.

\begin{lstlisting}[language=Python, numbers=none]
 evaluate.combine(["accuracy", "f1"])
\end{lstlisting}

\paragraph{Evaluator} \emph{Evaluate} also offers a higher level API called the Evaluator. Evaluator enables anyone to quickly evaluate a model on a task. Evaluator encapsulates task-specific pre- and post-processing and streamlines data preparation, model inference and metric computation. This makes the evaluation of any~\texttt{(model, dataset, metric)} triplet on a task seamless:~~\footnote{Currently text, token, and image classification as well as question-answering are supported with more coming soon.}

\begin{lstlisting}[language=Python, numbers=none]
 task = evaluator("text-classification")
 task.compute(model_or_pipeline=model,
              data=data, metric=metric)
\end{lstlisting}

Evaluator employs \emph{pipelines} from the Transformers library~\footnote{\href{https://huggingface.co/docs/transformers/main_classes/pipelines}{huggingface.co/docs/transformers/main\_classes/pipelines}} (or any other object with the same API) to carry out model inference. While evaluating downstream performance of the model, the Evaluator keeps track of the inference efficiency via metrics such as throughput and latency. This provides another dimension along which models can be compared, especially relevant in applied scenarios where inference times may be as crucial to a model's success as its performance on the core metrics. The Evaluator also supports (optional) confidence interval computations via bootstrapping on any metric.

\subsection{Documentation}

Recent years have seen several proposals for standardized documentation of both models~\cite{mitchell2019modelcards} and datasets~\cite{gebru2021datasheets}, arguing that this improves their accessibility as well as enabling a better understanding of their limitations and biases across different audiences. We have adopted this line of work within \textit{Evaluate} -- accompanying each evaluation module is a documentation card that describes the measurement, metric or comparison and how to use it. This card includes its intended use (i.e., whether it is specific to a task such as machine translation or a dataset such as SQuAD), its range, and code snippets that a user can copy within their application. These cards also contain a section on limitations and biases of the module, such as their applicability for certain languages (this is especially relevant for metrics such as \href{https://huggingface.co/spaces/evaluate-metric/bertscore}{BERTScore} and \href{https://huggingface.co/spaces/evaluate-metric/comet}{COMET}, which leverage pretrained models), the size of the models used to calculate them (e.g., GPT-2, the default model used for calculating \href{https://huggingface.co/spaces/evaluate-metric/mauve}{MAUVE}, is over 3 GB), and the fact that certain modules (e.g., \href{https://huggingface.co/spaces/evaluate-metric/perplexity}{perplexity}) are not comparable across different datasets when built from different models or preprocessing steps.

Our goal with these documentation cards is two-fold. On the one hand, we hope that they will \textit{educate} users regarding the scope and intention of different evaluation approaches, how they are calculated and how to interpret their values. On the other hand, we aim to \textit{improve best practices} in terms of evaluation approaches. This can be as simple as measuring \href{https://huggingface.co/spaces/evaluate-metric/f1}{F1 score} instead of relying simply on \href{https://huggingface.co/spaces/evaluate-metric/accuracy}{accuracy} for imbalanced datasets, but also preferring a more reproducible and systematic metric such as \href{https://huggingface.co/spaces/evaluate-metric/sacrebleu}{SacreBLEU} over a more variable one such as \href{https://huggingface.co/spaces/evaluate-metric/bleu}{BLEU}. We encourage the creators of new modules to write documentation cards to inform the community regarding the intended usages of their metric, measurement, or comparison; their possible limitations and biases; and to provide examples of best practices for using them.

\subsection{Community Contributions}
Since the code for metrics is stored in individual repositories on the Hugging Face Hub, anyone can add new metrics and load them with \emph{Evaluate} without needing to wait for reviews or approval. Any piece of evaluation code can be easily pushed to the Hugging Face Hub, which allows for sharing the exact same implementation with direct collaborators and the broader research community. These community metrics complement the canonical modules and are stored under the user's namespace. The \emph{Evaluate} library also includes a command line interface (CLI) to make community contributions more accessible.

\begin{lstlisting}[language=Python, numbers=none]
 evaluate-cli create "My awesome metric"
\end{lstlisting}
This command creates a repository on the Hub, clones it, populates it with a template and pushes it to the Hub. The user only needs to implement the metric logic, write a README containing the metric card, and push their changes to the Hub using Git.
We automatically provide live interaction widgets for each module, allowing users to develop a proper intuition for evaluation modules' usage, along with access to their documentation. Furthermore, our community discussion feature~\footnote{\href{https://huggingface.co/docs/hub/repositories-pull-requests-discussions}{huggingface.co/docs/hub/repositories-pull-requests-discussions}} allows members of the community to flag problematic evaluations or to ask for details regarding results, which model creators can then engage with. 

\begin{figure*}[t]
    \centering
    \includegraphics[width=\textwidth]{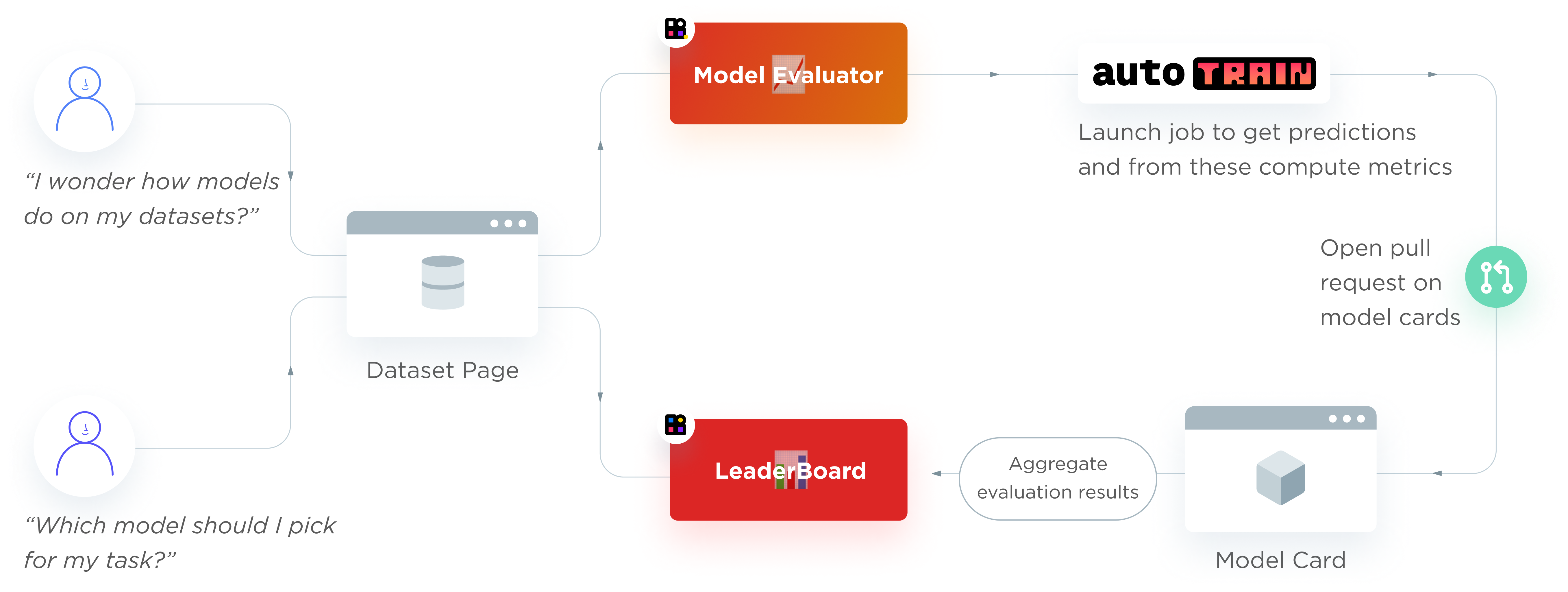}
    \caption{Evaluation on the Hub diagram}
    \label{fig:hub-diagram}
\end{figure*}

\section{Service: \textit{Evaluation on the Hub}}

The \emph{Evaluation on the Hub} platform extends the \textit{Evaluate} library to a free service model: anyone can evaluate any model on any dataset using any compatible metric, without requiring any code. This service utilizes models, datasets, and metrics standardized through the Hugging Face Hub. All evaluation results using this method are produced by the same pipeline with versioned implementations, and so are inherently reproducible. When a new model, dataset, or metric is produced, anyone can rerun the evaluation. As such, \emph{Evaluation on the Hub} facilitates large-scale evaluation of over 75,000 models and 11,000 datasets.

The service model further supports the goals of \textit{reproducibility} and \textit{centralization}. While the \emph{Evaluate} library can ensure that the metrics used are consistent, it cannot ensure that the model was trained and evaluated using a reproducible set of hyperparameters and data. Incorporating \emph{Evaluate} into a model hosting and training environment makes it possible to guarantee this consistency. Centralization also provides a further benefit of joining these metrics with model and data card documentation. 

\subsection{System architecture}

The system architecture is shown in Figure~\ref{fig:hub-diagram}. Upon submission, an evaluation job is triggered, which downloads the dataset and model(s) from the centralized Hub, computes metrics, and opens a pull request with the results. 

Evaluation jobs are configured through a simple interface~\footnote{\href{https://huggingface.co/spaces/autoevaluate/model-evaluator}{huggingface.co/spaces/autoevaluate/model-evaluator}} that specifies the task, dataset, metrics, and models to be evaluated. For each task, we compute a set of common metrics using the {\it Evaluate} library; users can also select additional metrics from the Hub~\footnote{\href{https://huggingface.co/metrics}{huggingface.co/metrics}} to be included in the evaluation. For many datasets on the Hub, we provide evaluation metadata that defines a default configuration for users to launch evaluation jobs with a single click. Users can also add evaluation metadata to their own datasets to provide one-click evaluations to the community. The interface for triggering an evaluation is shown in Figure~\ref{fig:hub-screenshots} (left).

We use AutoTrain~\footnote{\href{https://huggingface.co/autotrain}{huggingface.co/autotrain}}, Hugging Face's AutoML platform, to run evaluation jobs. The results from each evaluation are stored as metadata associated with model cards. The model predictions for each evaluation are also stored as dataset repositories on the Hub, enabling further analysis of, e.g., model errors.
    
\subsection{Documenting Evaluation}

The tool is permissioned so that model owners have the ability to select which evaluations they want to display with their model. This documentation is managed through a pull request system that allows owners to see evaluations that have been run. 
If a pull request is approved by the model owner, the results are added visibly to the model card as part of its documentation. However, all evaluation pull requests are public by default, so even if one is closed by model owners, members of the community can still see the scores.

\begin{figure*}[t]
    \centering
    \includegraphics[width=0.34\textwidth]{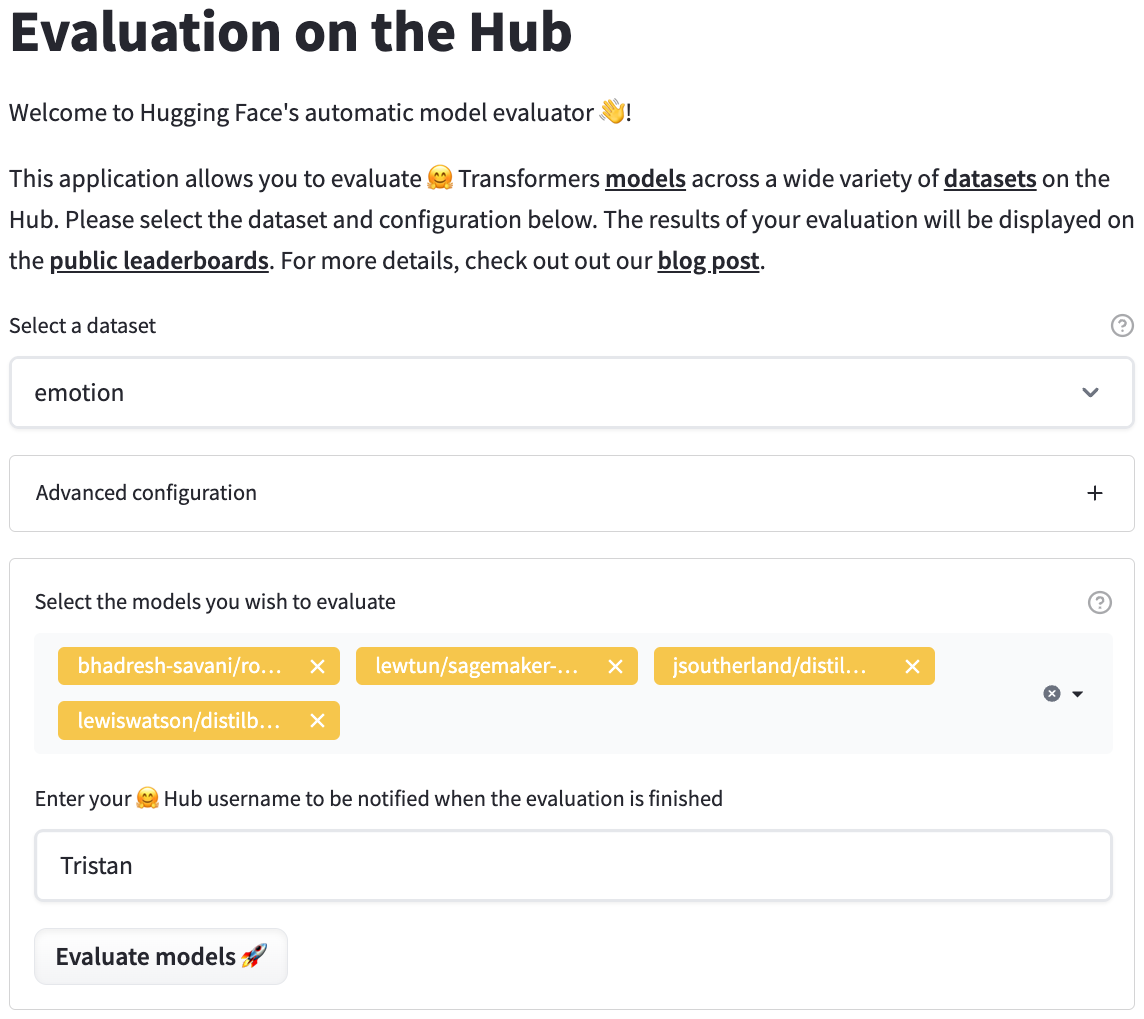}
    \unskip
    \hspace{0.15cm}
    \vrule
    \hspace{0.15cm}
    \includegraphics[width=0.63\textwidth]{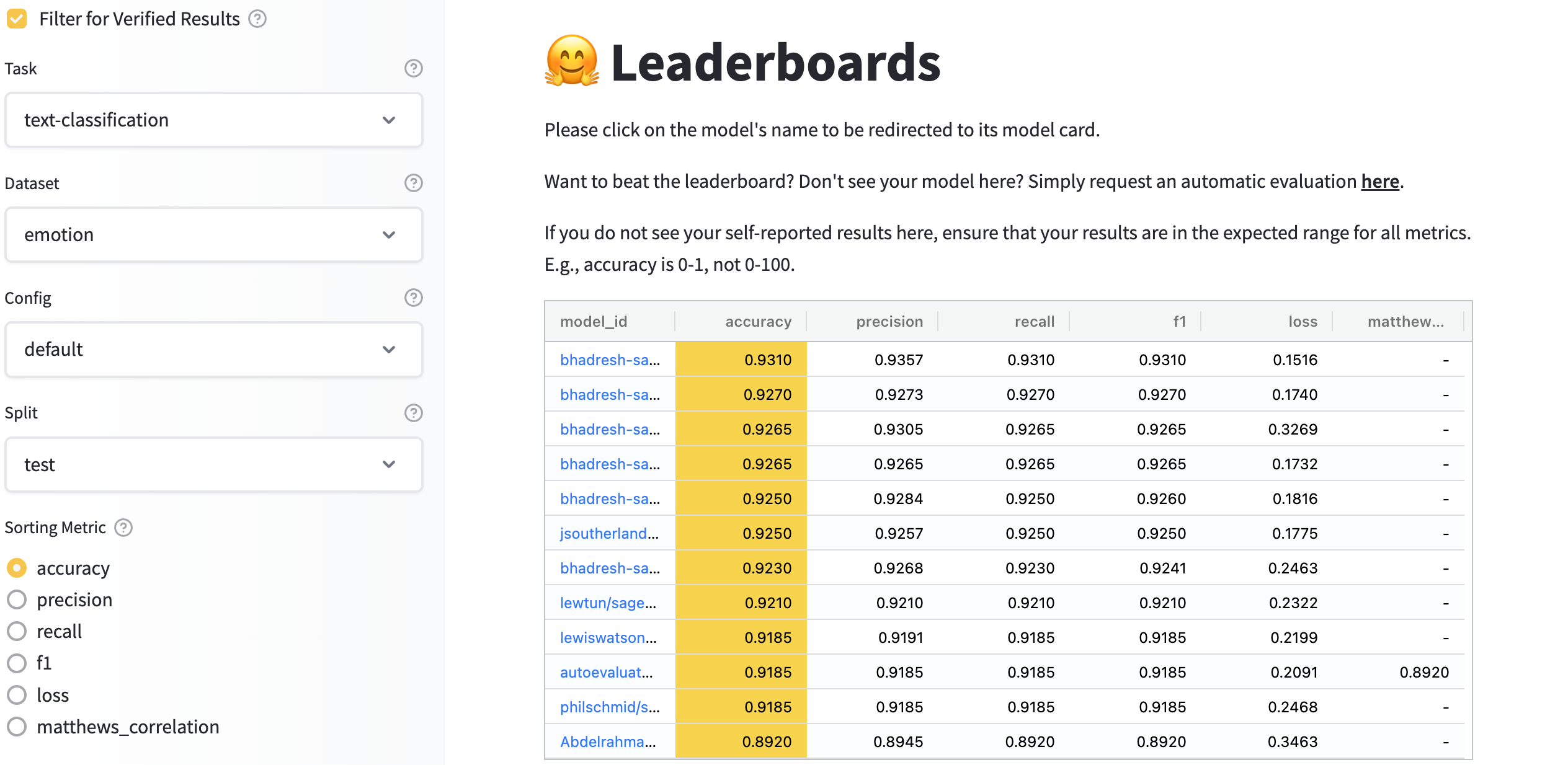}
    \caption{(left) Model Evaluator User Interface; (right) Leaderboards User Interface.}
    \label{fig:hub-screenshots}
\end{figure*}

Upon approval, the results become visible on an interactive Leaderboard~\footnote{\href{https://huggingface.co/spaces/autoevaluate/leaderboards}{huggingface.co/spaces/autoevaluate/leaderboards}} associated with the underlying dataset. We aggregate all model evaluations (both verified and self-reported) through these leaderboards that allow users to filter results across task and dataset. Models are ranked so that users can find the best scoring model for task X on dataset Y. The interface for model leaderboards is shown in Figure~\ref{fig:hub-screenshots} (left).

\section{Use Cases}

\emph{Evaluate} and \emph{Evaluation on the Hub} are already actively used by our community for a variety of tasks. There are many applications of these tools, and we highlight some of the most important use cases observed in practice.

\paragraph{Use case 1: Choosing the best model.}

If the task is known and the aim is to find an appropriate model, the Hub Leaderboard (which aggregates all the evaluation results for a dataset representative of that task) can act as a trusted source.  In case a particularly interesting model is not yet on the leaderboard, its evaluation can easily be triggered, directly from the Hub, and its results will automatically appear on the leaderboard, allowing it to be compared to previous models.

\paragraph{Use case 2: Reproducibility of results.}

If a new dataset is created, it can be uploaded directly to the Hub to trigger evaluation coverage on many models without needing any code. Researchers can trust in the reproducibility and consistency of this evaluation of these models on this dataset. Similarly, open-source implementations for measurements, metrics and comparisons can easily be shared and plugged into the Evaluator to enable reproducibility on a range of other model facets. If a paper does not report results for a given model on a dataset of interest, it can be evaluated and verified.

\paragraph{Use case 3: Deciding on deployment.}

When deciding on which variant of a model to deploy to production,  it is important to consider the broad performance of the model across multiple metrics. It may also be important to test on held-out test sets, and to measure the latency and throughput of a model. With the Evaluator, researchers can quickly evaluate on several datasets and also get the measured timing and latency information to make an informed decision.

\paragraph{Use case 4: Adding a new metric.}

When a new evaluation module (i.e., metric, measurement or comparison) is developed, it needs to be distributed for wider use. Historically, for
use-cases like Kaggle competitions, metrics are shared as code snippets, requiring participants to copy the evaluation code, which can be error-prone and inconvenient. With \emph{Evaluate}, anyone can create a new evaluate module---be it a metric, measurement, or comparison---alongside its documentation card with instructions.~\footnote{Example of a custom metric added by a community member: \href{https://huggingface.co/spaces/jordyvl/ece}{hf.co.co/spaces/jordyvl/ece}} Anybody with the access rights can then quickly use the module with the standard loading mechanism.

\section{Conclusion}

\emph{Evaluate} and \emph{Evaluation on the Hub} aim to facilitate better evaluation of machine learning data and models by improving reproducibility, centralization, and coverage of evaluation tools. \emph{Evaluate} is an open-source,
community-driven library that standardizes evaluation. \emph{Evaluation on the Hub} is a reproducible no-code alternative for evaluation across models, datasets, and metrics. We hope that this set of tools can help facilitate better best practices for model and data evaluation.

\section*{Ethical Issues and Limitations}
There are multiple aspects of model evaluation that we have not (yet) addressed but that remain important in the broader landscape of our community and the way ML is used in real-world settings. For instance, we have currently focused on metrics and measurements that have been developed and tested for high-resource languages such as English, and only cover a handful of metrics that explicitly support multilinguality. Similarly, while we strove to cover as many metrics as possible, most of our coverage is for text-based metrics, and we have yet to add as many metrics from other modalities, multi-modal metrics, or to provide as large a selection for measurements and comparisons. Furthermore, while we have documented the computational and memory requirements of our evaluation approaches via documentation cards, several metrics require downloading large models such as GPT-2, which can be inaccessible for users with slower Internet speeds or insufficient memory. Finally, we are still working towards a greater reproducibility of evaluation results, for instance by adding identifiers that will indicate which version of a metric and dataset was used for evaluating a model (in the case of code changes, for instance), allowing users to easily replicate results if needed. We will continue improving our tools to address these limitations and provide support for more uses cases.



\section*{Acknowledgements}

We thank the many contributors to the Hugging Face open source ecosystem. We thank Chunte Lee for designing the Evaluation on the Hub diagram in Figure~\ref{fig:hub-diagram}.

\bibliography{anthology,custom}

\begin{thebibliography}{45}
\expandafter\ifx\csname natexlab\endcsname\relax\def\natexlab#1{#1}\fi

\bibitem[{Akhbardeh et~al.(2021)Akhbardeh, Arkhangorodsky, Biesialska, Bojar,
  Chatterjee, Chaudhary, Costa-jussa, Espa{\~n}a-Bonet, Fan, Federmann,
  Freitag, Graham, Grundkiewicz, Haddow, Harter, Heafield, Homan, Huck,
  Amponsah-Kaakyire, Kasai, Khashabi, Knight, Kocmi, Koehn, Lourie, Monz,
  Morishita, Nagata, Nagesh, Nakazawa, Negri, Pal, Tapo, Turchi, Vydrin, and
  Zampieri}]{akhbardeh-etal-2021-findings}
Farhad Akhbardeh, Arkady Arkhangorodsky, Magdalena Biesialska, Ond{\v{r}}ej
  Bojar, Rajen Chatterjee, Vishrav Chaudhary, Marta~R. Costa-jussa, Cristina
  Espa{\~n}a-Bonet, Angela Fan, Christian Federmann, Markus Freitag, Yvette
  Graham, Roman Grundkiewicz, Barry Haddow, Leonie Harter, Kenneth Heafield,
  Christopher Homan, Matthias Huck, Kwabena Amponsah-Kaakyire, Jungo Kasai,
  Daniel Khashabi, Kevin Knight, Tom Kocmi, Philipp Koehn, Nicholas Lourie,
  Christof Monz, Makoto Morishita, Masaaki Nagata, Ajay Nagesh, Toshiaki
  Nakazawa, Matteo Negri, Santanu Pal, Allahsera~Auguste Tapo, Marco Turchi,
  Valentin Vydrin, and Marcos Zampieri. 2021.
\newblock \href {https://aclanthology.org/2021.wmt-1.1/} {Findings of the 2021
  conference on machine translation ({WMT}21)}.
\newblock \emph{WMT at EMNLP}.

\bibitem[{Ananthakrishnan et~al.(2007)Ananthakrishnan, Bhattacharyya,
  Sasikumar, and Shah}]{ananthakrishnan2007some}
R~Ananthakrishnan, Pushpak Bhattacharyya, M~Sasikumar, and Ritesh~M Shah. 2007.
\newblock \href {https://www.cse.iitb.ac.in/~pb/papers/icon07-bleu.pdf} {Some
  issues in automatic evaluation of {E}nglish-{H}indi {MT}: more blues for
  {BLEU}}.
\newblock \emph{ICON}, 64.

\bibitem[{Bajaj et~al.(2016)Bajaj, Campos, Craswell, Deng, Gao, Liu, Majumder,
  McNamara, Mitra, Nguyen et~al.}]{bajaj2016ms}
Payal Bajaj, Daniel Campos, Nick Craswell, Li~Deng, Jianfeng Gao, Xiaodong Liu,
  Rangan Majumder, Andrew McNamara, Bhaskar Mitra, Tri Nguyen, et~al. 2016.
\newblock \href {https://arxiv.org/abs/1611.09268} {{MS M}arco: A human
  generated machine reading comprehension dataset}.
\newblock \emph{arXiv preprint arXiv:1611.09268}.

\bibitem[{Bird et~al.(2009)Bird, Klein, and Loper}]{bird2009natural}
Steven Bird, Ewan Klein, and Edward Loper. 2009.
\newblock \href {https://www.nltk.org/book/} {\emph{Natural language processing
  with Python: analyzing text with the natural language toolkit}}.
\newblock " O'Reilly Media, Inc.".

\bibitem[{Blagec et~al.(2021)Blagec, Dorffner, Moradi, and
  Samwald}]{Blagec-et-al-2021}
Kathrin Blagec, Georg Dorffner, Milad Moradi, and Matthias Samwald. 2021.
\newblock \href {https://arxiv.org/abs/2008.02577} {A critical analysis of
  metrics used for measuring progress in artificial intelligence}.
\newblock \emph{arXiv pre-print}, 2008.02577:1--28.

\bibitem[{Bowman and Dahl(2021)}]{bowman2021will}
Samuel~R Bowman and George~E Dahl. 2021.
\newblock \href {https://arxiv.org/abs/2104.02145} {What will it take to fix
  benchmarking in natural language understanding?}
\newblock \emph{arXiv preprint arXiv:2104.02145}.

\bibitem[{Chan et~al.(2020)Chan, Fishman, Canny, Korattikara, and
  Guadarrama}]{chan2020rl}
Stephanie Chan, Sam Fishman, John Canny, Anoop Korattikara, and Sergio
  Guadarrama. 2020.
\newblock \href {https://openreview.net/pdf?id=SJlpYJBKvH} {Measuring the
  reliability of reinforcement learning algorithms}.
\newblock In \emph{International Conference on Learning Representations, Addis
  Ababa, Ethiopia}.

\bibitem[{Chollet et~al.(2015)}]{chollet2015keras}
Fran\c{c}ois Chollet et~al. 2015.
\newblock Keras.
\newblock \url{https://keras.io}.

\bibitem[{Coleman et~al.(2017)Coleman, Narayanan, Kang, Zhao, Zhang, Nardi,
  Bailis, Olukotun, R{\'e}, and Zaharia}]{coleman2017dawnbench}
Cody Coleman, Deepak Narayanan, Daniel Kang, Tian Zhao, Jian Zhang, Luigi
  Nardi, Peter Bailis, Kunle Olukotun, Chris R{\'e}, and Matei Zaharia. 2017.
\newblock \href
  {https://dawn.cs.stanford.edu/benchmark/papers/nips17-dawnbench.pdf}
  {Dawnbench: An end-to-end deep learning benchmark and competition}.
\newblock In \emph{NIPS ML Systems Workshop}.

\bibitem[{Detlefsen et~al.(2022)Detlefsen, Borovec, Schock, Harsh, Koker,
  Liello, Stancl, Quan, Grechkin, and Falcon}]{detlefsen22metrics}
Nicki~Skafte Detlefsen, Jiri Borovec, Justus Schock, Ananya Harsh, Teddy Koker,
  Luca~Di Liello, Daniel Stancl, Changsheng Quan, Maxim Grechkin, and William
  Falcon. 2022.
\newblock \href {https://github.com/Lightning-AI/metrics} {{TorchMetrics -
  Measuring Reproducibility in PyTorch}}.

\bibitem[{Dror et~al.(2019)Dror, Shlomov, and Reichart}]{dror2019deep}
Rotem Dror, Segev Shlomov, and Roi Reichart. 2019.
\newblock \href {https://aclanthology.org/P19-1266/} {Deep dominance-how to
  properly compare deep neural models}.
\newblock In \emph{Proceedings of the 57th Annual Meeting of the Association
  for Computational Linguistics}, pages 2773--2785.

\bibitem[{Ethayarajh and Jurafsky(2020)}]{ethayarajh2020utility}
Kawin Ethayarajh and Dan Jurafsky. 2020.
\newblock \href {https://arxiv.org/abs/2009.13888} {Utility is in the eye of
  the user: A critique of {NLP} leaderboards}.
\newblock \emph{arXiv preprint arXiv:2009.13888}.

\bibitem[{Gebru et~al.(2021)Gebru, Morgenstern, Vecchione, Vaughan, Wallach,
  Iii, and Crawford}]{gebru2021datasheets}
Timnit Gebru, Jamie Morgenstern, Briana Vecchione, Jennifer~Wortman Vaughan,
  Hanna Wallach, Hal~Daum{\'e} Iii, and Kate Crawford. 2021.
\newblock \href {https://arxiv.org/abs/1803.09010} {Datasheets for datasets}.
\newblock \emph{Communications of the ACM}, 64(12):86--92.

\bibitem[{Gehrmann et~al.(2021)Gehrmann, Adewumi, Aggarwal, Ammanamanchi,
  Aremu, Bosselut, Chandu, Clinciu, Das, Dhole, Du, Durmus, Du{\v{s}}ek,
  Emezue, Gangal, Garbacea, Hashimoto, Hou, Jernite, Jhamtani, Ji, Jolly, Kale,
  Kumar, Ladhak, Madaan, Maddela, Mahajan, Mahamood, Majumder, Martins,
  McMillan-Major, Mille, van Miltenburg, Nadeem, Narayan, Nikolaev,
  Niyongabo~Rubungo, Osei, Parikh, Perez-Beltrachini, Rao, Raunak, Rodriguez,
  Santhanam, Sedoc, Sellam, Shaikh, Shimorina, Sobrevilla~Cabezudo, Strobelt,
  Subramani, Xu, Yang, Yerukola, and Zhou}]{gehrmann-etal-2021-gem}
Sebastian Gehrmann, Tosin Adewumi, Karmanya Aggarwal, Pawan~Sasanka
  Ammanamanchi, Anuoluwapo Aremu, Antoine Bosselut, Khyathi~Raghavi Chandu,
  Miruna-Adriana Clinciu, Dipanjan Das, Kaustubh Dhole, Wanyu Du, Esin Durmus,
  Ond{\v{r}}ej Du{\v{s}}ek, Chris~Chinenye Emezue, Varun Gangal, Cristina
  Garbacea, Tatsunori Hashimoto, Yufang Hou, Yacine Jernite, Harsh Jhamtani,
  Yangfeng Ji, Shailza Jolly, Mihir Kale, Dhruv Kumar, Faisal Ladhak, Aman
  Madaan, Mounica Maddela, Khyati Mahajan, Saad Mahamood, Bodhisattwa~Prasad
  Majumder, Pedro~Henrique Martins, Angelina McMillan-Major, Simon Mille, Emiel
  van Miltenburg, Moin Nadeem, Shashi Narayan, Vitaly Nikolaev, Andre
  Niyongabo~Rubungo, Salomey Osei, Ankur Parikh, Laura Perez-Beltrachini,
  Niranjan~Ramesh Rao, Vikas Raunak, Juan~Diego Rodriguez, Sashank Santhanam,
  Jo{\~a}o Sedoc, Thibault Sellam, Samira Shaikh, Anastasia Shimorina,
  Marco~Antonio Sobrevilla~Cabezudo, Hendrik Strobelt, Nishant Subramani, Wei
  Xu, Diyi Yang, Akhila Yerukola, and Jiawei Zhou. 2021.
\newblock \href {https://doi.org/10.18653/v1/2021.gem-1.10} {The {GEM}
  benchmark: Natural language generation, its evaluation and metrics}.
\newblock In \emph{Proceedings of the 1st Workshop on Natural Language
  Generation, Evaluation, and Metrics (GEM 2021)}, pages 96--120, Online.
  Association for Computational Linguistics.

\bibitem[{Goel et~al.(2021)Goel, Rajani, Vig, Tan, Wu, Zheng, Xiong, Bansal,
  and R{\'e}}]{goel2021robustness}
Karan Goel, Nazneen Rajani, Jesse Vig, Samson Tan, Jason Wu, Stephan Zheng,
  Caiming Xiong, Mohit Bansal, and Christopher R{\'e}. 2021.
\newblock \href {https://aclanthology.org/2021.naacl-demos.6.pdf} {Robustness
  gym: Unifying the {NLP} evaluation landscape}.
\newblock \emph{arXiv preprint arXiv:2101.04840}.

\bibitem[{Hagberg et~al.(2008)Hagberg, Swart, and
  S~Chult}]{hagberg2008exploring}
Aric Hagberg, Pieter Swart, and Daniel S~Chult. 2008.
\newblock \href {https://www.osti.gov/biblio/960616} {Exploring network
  structure, dynamics, and function using networkx}.
\newblock Technical report, Los Alamos National Lab.(LANL), Los Alamos, NM
  (United States).

\bibitem[{Hutchinson et~al.(2022)Hutchinson, Rostamzadeh, Greer, Heller, and
  Prabhakaran}]{hutchinson2022evaluation}
Ben Hutchinson, Negar Rostamzadeh, Christina Greer, Katherine Heller, and
  Vinodkumar Prabhakaran. 2022.
\newblock \href {https://dl.acm.org/doi/fullHtml/10.1145/3531146.3533233}
  {Evaluation gaps in machine learning practice}.
\newblock \emph{arXiv preprint arXiv:2205.05256}.

\bibitem[{Kardas et~al.(2020)Kardas, Czapla, Stenetorp, Ruder, Riedel, Taylor,
  and Stojnic}]{kardas-etal-2020-axcell}
Marcin Kardas, Piotr Czapla, Pontus Stenetorp, Sebastian Ruder, Sebastian
  Riedel, Ross Taylor, and Robert Stojnic. 2020.
\newblock \href {https://aclanthology.org/2020.emnlp-main.692/} {{AxCell}:
  Automatic extraction of results from machine learning papers}.
\newblock \emph{EMNLP}.

\bibitem[{Kiela et~al.(2021)Kiela, Bartolo, Nie, Kaushik, Geiger, Wu, Vidgen,
  Prasad, Singh, Ringshia et~al.}]{kiela2021dynabench}
Douwe Kiela, Max Bartolo, Yixin Nie, Divyansh Kaushik, Atticus Geiger,
  Zhengxuan Wu, Bertie Vidgen, Grusha Prasad, Amanpreet Singh, Pratik Ringshia,
  et~al. 2021.
\newblock \href {https://aclanthology.org/2021.naacl-main.324/} {Dynabench:
  Rethinking benchmarking in {NLP}}.
\newblock \emph{arXiv preprint arXiv:2104.14337}.

\bibitem[{Lhoest et~al.(2021)Lhoest, del Moral, Jernite, Thakur, von Platen,
  Patil, Chaumond, Drame, Plu, Tunstall et~al.}]{lhoest2021datasets}
Quentin Lhoest, Albert~Villanova del Moral, Yacine Jernite, Abhishek Thakur,
  Patrick von Platen, Suraj Patil, Julien Chaumond, Mariama Drame, Julien Plu,
  Lewis Tunstall, et~al. 2021.
\newblock \href {https://aclanthology.org/2021.emnlp-demo.21/} {Datasets: A
  community library for natural language processing}.
\newblock \emph{arXiv preprint arXiv:2109.02846}.

\bibitem[{Liu et~al.(2021)Liu, Fu, Xiao, Yuan, Chang, Dai, Liu, Ye, and
  Neubig}]{liu2021explainaboard}
Pengfei Liu, Jinlan Fu, Yang Xiao, Weizhe Yuan, Shuaicheng Chang, Junqi Dai,
  Yixin Liu, Zihuiwen Ye, and Graham Neubig. 2021.
\newblock \href {https://aclanthology.org/2021.acl-demo.34/} {{EXPLAINABOARD:
  An Explainable Leaderboard for NLP}}.
\newblock \emph{arXiv preprint arXiv:2104.06387}.

\bibitem[{Luccioni et~al.(2022)Luccioni, Corry, Sridharan, Ananny, Schultz, and
  Crawford}]{luccioni2022framework}
Alexandra~Sasha Luccioni, Frances Corry, Hamsini Sridharan, Mike Ananny, Jason
  Schultz, and Kate Crawford. 2022.
\newblock \href {https://dl.acm.org/doi/abs/10.1145/3531146.3533086} {A
  framework for deprecating datasets: Standardizing documentation,
  identification, and communication}.
\newblock In \emph{2022 ACM Conference on Fairness, Accountability, and
  Transparency}, pages 199--212.

\bibitem[{Ma et~al.(2021)Ma, Ethayarajh, Thrush, Jain, Wu, Jia, Potts,
  Williams, and Kiela}]{ma2021dynaboard}
Zhiyi Ma, Kawin Ethayarajh, Tristan Thrush, Somya Jain, Ledell Wu, Robin Jia,
  Christopher Potts, Adina Williams, and Douwe Kiela. 2021.
\newblock \href {https://dynabench.org/dynaboard.pdf} {Dynaboard: An
  evaluation-as-a-service platform for holistic next-generation benchmarking}.
\newblock \emph{Advances in Neural Information Processing Systems},
  34:10351--10367.

\bibitem[{Marie et~al.(2021)Marie, Fujita, and Rubino}]{marie2021scientific}
Benjamin Marie, Atsushi Fujita, and Raphael Rubino. 2021.
\newblock \href {https://aclanthology.org/2021.acl-long.566.pdf} {Scientific
  credibility of machine translation research: A meta-evaluation of 769
  papers}.
\newblock \emph{arXiv preprint arXiv:2106.15195}.

\bibitem[{Min et~al.(2021)Min, Boyd-Graber, Alberti, Chen, Choi, Collins, Guu,
  Hajishirzi, Lee, Palomaki, Raffel, Roberts, Kwiatkowski, Lewis, Wu,
  K\"uttler, Liu, Minervini, Stenetorp, Riedel, Yang, Seo, Izacard, Petroni,
  Hosseini, Cao, Grave, Yamada, Shimaoka, Suzuki, Miyawaki, Sato, Takahashi,
  Suzuki, Fajcik, Docekal, Ondrej, Smrz, Cheng, Shen, Liu, He, Chen, Gao, Oguz,
  Chen, Karpukhin, Peshterliev, Okhonko, Schlichtkrull, Gupta, Mehdad, and
  Yih}]{efficientqa2021}
Sewon Min, Jordan Boyd-Graber, Chris Alberti, Danqi Chen, Eunsol Choi, Michael
  Collins, Kelvin Guu, Hannaneh Hajishirzi, Kenton Lee, Jennimaria Palomaki,
  Colin Raffel, Adam Roberts, Tom Kwiatkowski, Patrick Lewis, Yuxiang Wu,
  Heinrich K\"uttler, Linqing Liu, Pasquale Minervini, Pontus Stenetorp,
  Sebastian Riedel, Sohee Yang, Minjoon Seo, Gautier Izacard, Fabio Petroni,
  Lucas Hosseini, Nicola~De Cao, Edouard Grave, Ikuya Yamada, Sonse Shimaoka,
  Masatoshi Suzuki, Shumpei Miyawaki, Shun Sato, Ryo Takahashi, Jun Suzuki,
  Martin Fajcik, Martin Docekal, Karel Ondrej, Pavel Smrz, Hao Cheng, Yelong
  Shen, Xiaodong Liu, Pengcheng He, Weizhu Chen, Jianfeng Gao, Barlas Oguz,
  Xilun Chen, Vladimir Karpukhin, Stan Peshterliev, Dmytro Okhonko, Michael
  Schlichtkrull, Sonal Gupta, Yashar Mehdad, and Wen-tau Yih. 2021.
\newblock \href {https://proceedings.mlr.press/v133/min21a.html} {{NeurIPS 2020
  EfficientQA} competition: Systems, analyses and lessons learned}.
\newblock In \emph{Proceedings of the NeurIPS 2020 Competition and
  Demonstration Track}, volume 133 of \emph{Proceedings of Machine Learning
  Research}, pages 86--111. PMLR.

\bibitem[{Mitchell et~al.(2019)Mitchell, Wu, Zaldivar, Barnes, Vasserman,
  Hutchinson, Spitzer, Raji, and Gebru}]{mitchell2019modelcards}
Margaret Mitchell, Simone Wu, Andrew Zaldivar, Parker Barnes, Lucy Vasserman,
  Ben Hutchinson, Elena Spitzer, Inioluwa~Deborah Raji, and Timnit Gebru. 2019.
\newblock \href {https://doi.org/10.1145/3287560.3287596} {Model cards for
  model reporting}.
\newblock In \emph{Proceedings of the Conference on Fairness, Accountability,
  and Transparency}, FAT* '19, page 220–229, New York, NY, USA. Association
  for Computing Machinery.

\bibitem[{Obukhov et~al.(2020)Obukhov, Seitzer, Wu, Zhydenko, Kyl, and
  Lin}]{obukhov2020torchfidelity}
Anton Obukhov, Maximilian Seitzer, Po-Wei Wu, Semen Zhydenko, Jonathan Kyl, and
  Elvis Yu-Jing Lin. 2020.
\newblock \href {https://doi.org/10.5281/zenodo.4957738} {{High-fidelity
  performance metrics for generative models in PyTorch}}.
\newblock Version: 0.3.0, DOI: 10.5281/zenodo.4957738.

\bibitem[{Palotti et~al.(2019)Palotti, Scells, and Zuccon}]{palotti2019}
Joao Palotti, Harrisen Scells, and Guido Zuccon. 2019.
\newblock \href {https://dl.acm.org/doi/10.1145/3331184.3331399} {Trectools: an
  open-source {P}ython library for information retrieval practitioners involved
  in {TREC}-like campaigns}.
\newblock SIGIR'19. ACM.

\bibitem[{Papineni et~al.(2002)Papineni, Roukos, Ward, and
  Zhu}]{papineni2002bleu}
Kishore Papineni, Salim Roukos, Todd Ward, and Wei-Jing Zhu. 2002.
\newblock \href {https://aclanthology.org/P02-1040/} {{BLEU}: a method for
  automatic evaluation of machine translation}.
\newblock In \emph{Proceedings of the 40th annual meeting of the Association
  for Computational Linguistics}, pages 311--318.

\bibitem[{Paszke et~al.(2019)Paszke, Gross, Massa, Lerer, Bradbury, Chanan,
  Killeen, Lin, Gimelshein, Antiga et~al.}]{paszke2019pytorch}
Adam Paszke, Sam Gross, Francisco Massa, Adam Lerer, James Bradbury, Gregory
  Chanan, Trevor Killeen, Zeming Lin, Natalia Gimelshein, Luca Antiga, et~al.
  2019.
\newblock \href
  {https://proceedings.neurips.cc/paper/2019/file/bdbca288fee7f92f2bfa9f7012727740-Paper.pdf}
  {Pytorch: An imperative style, high-performance deep learning library}.
\newblock \emph{Advances in neural information processing systems}, 32.

\bibitem[{Pavao et~al.(2022)Pavao, Guyon, Letournel, Bar{\'o}, Escalante,
  Escalera, Thomas, and Xu}]{pavao2022codalab}
Adrien Pavao, Isabelle Guyon, Anne-Catherine Letournel, Xavier Bar{\'o}, Hugo
  Escalante, Sergio Escalera, Tyler Thomas, and Zhen Xu. 2022.
\newblock \href {https://tel.archives-ouvertes.fr/LISN-AO/hal-03629462v1}
  {\emph{CodaLab Competitions: An open source platform to organize scientific
  challenges}}.
\newblock Ph.D. thesis, Universit{\'e} Paris-Saclay, France.

\bibitem[{Pedregosa et~al.(2011)Pedregosa, Varoquaux, Gramfort, Michel,
  Thirion, Grisel, Blondel, Prettenhofer, Weiss, Dubourg
  et~al.}]{pedregosa2011scikit}
Fabian Pedregosa, Ga{\"e}l Varoquaux, Alexandre Gramfort, Vincent Michel,
  Bertrand Thirion, Olivier Grisel, Mathieu Blondel, Peter Prettenhofer, Ron
  Weiss, Vincent Dubourg, et~al. 2011.
\newblock \href {https://jmlr.org/papers/v12/pedregosa11a.html} {Scikit-learn:
  {Ma}chine learning in {P}ython}.
\newblock \emph{the Journal of machine Learning research}, 12:2825--2830.

\bibitem[{Pineau et~al.(2021)Pineau, Vincent-Lamarre, Sinha, Larivi{\`e}re,
  Beygelzimer, d’Alch{\'e} Buc, Fox, and Larochelle}]{pineau2021improving}
Joelle Pineau, Philippe Vincent-Lamarre, Koustuv Sinha, Vincent Larivi{\`e}re,
  Alina Beygelzimer, Florence d’Alch{\'e} Buc, Emily Fox, and Hugo
  Larochelle. 2021.
\newblock \href {https://www.jmlr.org/papers/volume22/20-303/20-303.pdf}
  {Improving reproducibility in machine learning research: a report from the
  {NeurIPS} 2019 reproducibility program}.
\newblock \emph{Journal of Machine Learning Research}, 22.

\bibitem[{Post(2018)}]{post2018call}
Matt Post. 2018.
\newblock \href {https://aclanthology.org/W18-6319.pdf} {A call for clarity in
  reporting {BLEU} scores}.
\newblock \emph{arXiv preprint arXiv:1804.08771}.

\bibitem[{Qian et~al.(2022)Qian, Ross, Fernandes, Smith, Kiela, and
  Williams}]{qian2022perturbation}
Rebecca Qian, Candace Ross, Jude Fernandes, Eric Smith, Douwe Kiela, and Adina
  Williams. 2022.
\newblock \href {https://arxiv.org/abs/2205.12586} {Perturbation augmentation
  for fairer {NLP}}.
\newblock \emph{arXiv preprint arXiv:2205.12586}.

\bibitem[{Raff(2019)}]{raff2019step}
Edward Raff. 2019.
\newblock \href {https://arxiv.org/abs/1909.06674} {A step toward quantifying
  independently reproducible machine learning research}.
\newblock \emph{Advances in Neural Information Processing Systems}, 32.

\bibitem[{Raji et~al.(2021)Raji, Bender, Paullada, Denton, and
  Hanna}]{raji2021ai}
Inioluwa~Deborah Raji, Emily~M Bender, Amandalynne Paullada, Emily Denton, and
  Alex Hanna. 2021.
\newblock \href {https://arxiv.org/abs/2111.15366} {{AI} and the everything in
  the whole wide world benchmark}.
\newblock \emph{arXiv preprint arXiv:2111.15366}.

\bibitem[{Reddi et~al.(2020)Reddi, Cheng, Kanter, Mattson, Schmuelling, Wu,
  Anderson, Breughe, Charlebois, Chou, Chukka, Coleman, Davis, Deng, Diamos,
  Duke, Fick, Gardner, Hubara, Idgunji, Jablin, Jiao, John, Kanwar, Lee, Liao,
  Lokhmotov, Massa, Meng, Micikevicius, Osborne, Pekhimenko, Rajan, Sequeira,
  Sirasao, Sun, Tang, Thomson, Wei, Wu, Xu, Yamada, Yu, Yuan, Zhong, Zhang, and
  Zhou}]{reddi-2020-mlperf}
Vijay~Janapa Reddi, Christine Cheng, David Kanter, Peter Mattson, Guenther
  Schmuelling, Carole{-}Jean Wu, Brian Anderson, Maximilien Breughe, Mark
  Charlebois, William Chou, Ramesh Chukka, Cody Coleman, Sam Davis, Pan Deng,
  Greg Diamos, Jared Duke, Dave Fick, J.~Scott Gardner, Itay Hubara, Sachin
  Idgunji, Thomas~B. Jablin, Jeff Jiao, Tom~St. John, Pankaj Kanwar, David Lee,
  Jeffery Liao, Anton Lokhmotov, Francisco Massa, Peng Meng, Paulius
  Micikevicius, Colin Osborne, Gennady Pekhimenko, Arun Tejusve~Raghunath
  Rajan, Dilip Sequeira, Ashish Sirasao, Fei Sun, Hanlin Tang, Michael Thomson,
  Frank Wei, Ephrem Wu, Lingjie Xu, Koichi Yamada, Bing Yu, George Yuan, Aaron
  Zhong, Peizhao Zhang, and Yuchen Zhou. 2020.
\newblock \href {https://arxiv.org/abs/1911.02549} {{MLPerf Inference
  Benchmark}}.
\newblock \emph{ISCA}.

\bibitem[{Seabold and Perktold(2010)}]{seabold2010statsmodels}
Skipper Seabold and Josef Perktold. 2010.
\newblock statsmodels: Econometric and statistical modeling with python.
\newblock In \emph{9th Python in Science Conference}.

\bibitem[{Van~der Walt et~al.(2014)Van~der Walt, Sch{\"o}nberger,
  Nunez-Iglesias, Boulogne, Warner, Yager, Gouillart, and Yu}]{van2014scikit}
Stefan Van~der Walt, Johannes~L Sch{\"o}nberger, Juan Nunez-Iglesias,
  Fran{\c{c}}ois Boulogne, Joshua~D Warner, Neil Yager, Emmanuelle Gouillart,
  and Tony Yu. 2014.
\newblock \href
  {https://peerj.com/articles/453/?report=reader&utm_source=TrendMD&utm_campaign=PeerJ_TrendMD_1&utm_medium=TrendMD}
  {scikit-image: image processing in python}.
\newblock \emph{PeerJ}, 2:e453.

\bibitem[{Virtanen et~al.(2020)Virtanen, Gommers, Oliphant, Haberland, Reddy,
  Cournapeau, Burovski, Peterson, Weckesser, Bright et~al.}]{virtanen2020scipy}
Pauli Virtanen, Ralf Gommers, Travis~E Oliphant, Matt Haberland, Tyler Reddy,
  David Cournapeau, Evgeni Burovski, Pearu Peterson, Warren Weckesser, Jonathan
  Bright, et~al. 2020.
\newblock \href {https://www.nature.com/articles/s41592-019-0686-2} {Scipy 1.0:
  fundamental algorithms for scientific computing in python}.
\newblock \emph{Nature methods}, 17(3):261--272.

\bibitem[{Wang et~al.(2019)Wang, Pruksachatkun, Nangia, Singh, Michael, Hill,
  Levy, and Bowman}]{wang2019superglue}
Alex Wang, Yada Pruksachatkun, Nikita Nangia, Amanpreet Singh, Julian Michael,
  Felix Hill, Omer Levy, and Samuel Bowman. 2019.
\newblock \href
  {https://papers.nips.cc/paper/2019/hash/4496bf24afe7fab6f046bf4923da8de6-Abstract.html}
  {Super{GLUE}: A stickier benchmark for general-purpose language understanding
  systems}.
\newblock \emph{Advances in neural information processing systems}, 32.

\bibitem[{Wang et~al.(2018)Wang, Singh, Michael, Hill, Levy, and
  Bowman}]{wang2018glue}
Alex Wang, Amanpreet Singh, Julian Michael, Felix Hill, Omer Levy, and Samuel~R
  Bowman. 2018.
\newblock \href {https://aclanthology.org/W18-5446/} {{GLUE: A multi-task
  benchmark and analysis platform for natural language understanding}}.
\newblock \emph{arXiv preprint arXiv:1804.07461}.

\bibitem[{Yadav et~al.(2019)Yadav, Jain, Agrawal, Chattopadhyay, Singh, Jain,
  Singh, Lee, and Batra}]{EvalAI}
Deshraj Yadav, Rishabh Jain, Harsh Agrawal, Prithvijit Chattopadhyay, Taranjeet
  Singh, Akash Jain, Shiv~Baran Singh, Stefan Lee, and Dhruv Batra. 2019.
\newblock \href {https://arxiv.org/abs/1902.03570} {{EvalAI: Towards Better
  Evaluation Systems for AI Agents}}.
\newblock \emph{arXiv preprint arXiv:1902.03570}.

\bibitem[{Yuan et~al.(2021)Yuan, Neubig, and Liu}]{yuan2021bartscore}
Weizhe Yuan, Graham Neubig, and Pengfei Liu. 2021.
\newblock \href
  {https://proceedings.neurips.cc/paper/2021/file/e4d2b6e6fdeca3e60e0f1a62fee3d9dd-Paper.pdf}
  {Bartscore: Evaluating generated text as text generation}.
\newblock In \emph{Advances in Neural Information Processing Systems},
  volume~34, pages 27263--27277. Curran Associates, Inc.

\end{thebibliography}
\bibliographystyle{acl_natbib}

\end{document}